\documentclass{article}
\newcommand{\mycomment}[1]{}

\usepackage[final]{timeseries_workshop}

\usepackage[utf8]{inputenc} 
\usepackage[T1]{fontenc}    
\usepackage{hyperref}       
\usepackage{url}            
\usepackage{booktabs}       
\usepackage{amsfonts}       
\usepackage{nicefrac}       
\usepackage{microtype}      
\usepackage{xcolor}         
\usepackage{multirow}
\usepackage[pdftex]{graphicx}
\usepackage{amsmath} 

\title{Hierarchical Time Series Forecasting Via Latent Mean Encoding}

\author{%
  Alessandro Salatiello \\
  Zalando SE \\
  \texttt{alessandro.salatiello@uni-tuebingen.de}
  \And
  Stefan Birr \\
  Zalando SE \\
  \texttt{stefan.birr@zalando.de}
  \And
  Manuel Kunz \\
  Zalando SE \\
  \texttt{manuel.kunz@zalando.de}
}

\begin{document}

\maketitle

\begin{abstract}
Coherently forecasting the behaviour of a target variable across both coarse and fine temporal scales is crucial for profit-optimized decision-making in several business applications, and remains an open research problem in temporal hierarchical forecasting. 
Here, we propose a new hierarchical architecture that tackles this problem by leveraging modules that specialize in forecasting the different temporal aggregation levels of interest.
The architecture, which learns to encode the average behaviour of the target variable within its hidden layers, makes accurate and coherent forecasts across the target temporal hierarchies. We validate our architecture on the challenging, real-world M5 dataset and show that it outperforms established methods, such as the TSMixer model.
\end{abstract}
\section{Introduction}
Hierarchical forecasting studies methods to ensure coherent forecasts across the inherent hierarchies in data \citep{hyndman2011optimal}. It is a growing research field, particularly relevant in domains such as e-commerce \citep{rangapuram2023coherent,sprangers2024hierarchical} and electricity demand forecasting \citep{taieb2017coherent}. In these domains, cross-sectional hierarchies, such as geographical hierarchies or product category trees in retail demand forecasting \citep{kunz2023deep}, are naturally present in the data. Similarly, temporal hierarchies arise from the need to provide coherent forecasts on multiple time resolutions \citep{taieb2017sparse}. In retail demand forecasting, for example, high-resolution forecasts are required to steer inventory on a daily or weekly frequency. In contrast, strategic planning and finance need forecasts on a quarterly level for geographical regions or product categories \citep{sprangers2024hierarchical}. Although structural and temporal hierarchies can be intertwined \citep{kourentzes2019cross}, here we focus on providing coherent forecasts along the temporal dimension. This problem is often addressed by training independent, specialized forecasting models for each frequency of interest, followed by a post-processing reconciliation step to ensure coherence \citep{kourentzes2014improving}. Alternatively, established methods \citep{athanasopoulos2017forecasting} avoid the post-processing step by using a bottom-up approach that only requires training a single model at the observed, fine-grained frequency. With this work, we provide a novel approach to coherent temporal hierarchical forecasting based on using a single hierarchical neural network model with frequency-specific modules. Importantly, the network --- trained end-to-end on a state-of-the-art deep learning architecture --- does not require post-processing steps and outperforms established forecasting models on real-world data.
\section{Methods}
\newcommand{\minus}{{\scriptscriptstyle -}}
\newcommand{\plus}{{\scriptscriptstyle +}}
\begin{figure}[!b]
  \centering
  \includegraphics[width=\linewidth]{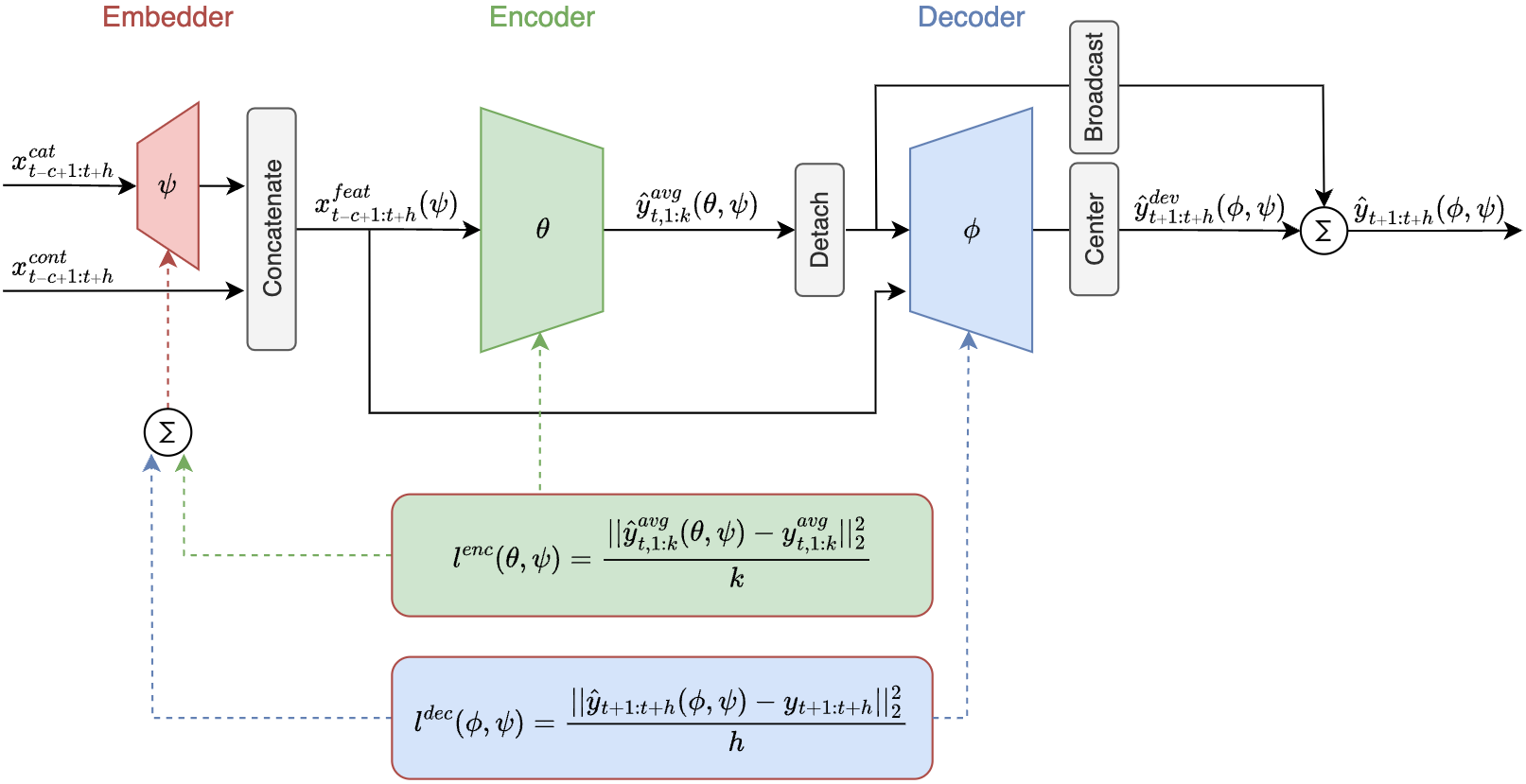}
  \caption{The proposed architecture. The architecture forecasts $h$ future values of a target variable $y$ using $c$ past and $h$ future values of categorical $x^{cat}$ and continuous $x^{cont}$ covariates. The architecture comprises three main modules — the embedder, the encoder, and the decoder modules — as well as a non-trainable readout layer. The embedder module learns a continuous representation of the categorical variables. The encoder module learns to predict $\smash{\hat{y}_{t,1:k}^{avg}(\theta,\psi)}$: the average values of the target variable over $k$ consecutive bins of size $w$. The decoder learns to predict $\hat{y}_{t+1:t \plus h}^{dev}(\phi,\psi)$: the deviations from the average values. Finally, a non-trainable readout layer computes the architecture output $\hat{y}_{t \plus 1:t \plus h}^{}(\phi,\psi)$ by summing aligned representations of the averages and deviations. The architecture is trained by minimizing two loss functions: the encoder loss $l^{enc}(\theta,\psi)$, which optimizes the encoder and the embedder, and the decoder loss $l^{dec}(\phi,\psi)$, which optimizes the decoder and the embedder.}
  \label{fig:architecture}
\end{figure}

In this work, we set out to develop an efficient architecture for coherent time series forecasting at multiple temporal aggregation levels. More formally, we aim to address the problem of simultaneously and coherently forecasting the future behaviour of a target variable $y(t)$ at its original base sampling frequency and at a coarser frequency over a forecasting horizon of $h$ samples. Specifically, we are interested in the coarse-grained time series $y(t)^{avg}$ obtained from the fine-grained one $y(t)$ by computing non-overlapping averages over $k$ bins of $w$ samples. Thus, in summary, our architecture should be able to forecast $y_{t \plus 1:t \plus h}:=\{y_{t+i}\}_{i=1}^{h}$ and $y^{avg}_{t,1:k}:=\{y^{avg}_{t,j}\}_{j=1}^{k}$ with $k=\frac{h}{w}$,
respecting the following constraint:
\begin{equation}
    y_{t,j}^{avg} = \frac{1}{w}\sum_{i=1}^{w}y_{t \plus i \plus w(j \minus 1)} 
\end{equation}
The predictions can be based on the past and future values of both categorical covariates $x^{cat}_{t \minus c \plus 1:t+h}$ and continuous covariates $x^{cont}_{t \minus c \plus 1:t \plus h}$, over a window extending $c$ samples into the past and $h$ samples into the future. Importantly, the covariates can include the past true values of the target variable $y_{t \minus c \plus 1:t}$ as well as static variables. 

\subsection{The encoder-decoder architecture}
 The architecture we developed to tackle this problem is represented in Fig.\ref{fig:architecture}. The architecture consists of three trainable modules — the embedder $b(z;\psi)$, the encoder $e(z;\theta)$, and the decoder $d(z;\phi)$ modules — which work synergistically to solve the hierarchical forecasting problem. Indeed, our design encourages the encoder module to specialize in forecasting the coarse-grained time series while the decoder module specializes in forecasting the fine-grained time series. Meanwhile, the embedder module specializes in learning continuous representations of the categorical covariates that are useful for forecasting both the coarse and fine-grained behaviour of the target variable.
 
 More precisely, the embedder module embeds the categorical variables $x^{cat}_{t \minus c \plus 1:t \plus h}$ into a dense, multidimensional continuous space; these are then concatenated with the continuous variables into a tensor $x^{feat}_{t \minus c \plus 1:t \plus h}(\psi)$ and forwarded to both the encoder and decoder modules. The encoder module learns to predict the average values of the target variable over $k$ consecutive bins of size $w$: $\hat{y}^{avg}_{t,1:k}(\theta,\psi)$. These average values are then forwarded to the decoder, which can, in turn, learn to predict the deviations from the average values $\hat{y}_{t \plus 1:t \plus h}^{dev}(\phi,\psi)$. This is enforced by postprocessing the decoder output with a centering module that subtracts the average values the decoder outputs over the $k$ bins of size $w$. 
 
 Finally, a non-trainable readout layer computes the architecture output $\hat{y}_{t \plus 1:t \plus h}^{}(\phi,\psi)$ by summing aligned representations of the averages and the deviations according to:
\begin{equation}
    \hat{y}_{t \plus 1:t \plus h}^{}(\phi,\psi) = \mathbf{S}_{w, k}\hat{y}^{avg}_{t,1:k}(\theta,\psi) + \hat{y}_{t \plus 1:t \plus h}^{dev}(\phi,\psi),
\end{equation}
with $\mathbf{S}_{w, k}=(s_{i,j})_{ 1 \leq i \leq h, \, 1 \leq j \leq k}$ defined by:
\begin{equation}
s_{i,j} = 
\begin{cases} 
1 & \text{if} \ (j-1)w < i \leq jw, \\
0 & \text{otherwise}.
\end{cases}
\end{equation}

 
 The architecture is trained by minimizing two loss functions. The encoder loss, which is used to optimize the encoder and the embedder modules:
 \begin{equation}
      l^{enc}(\theta, \psi) = \frac{|| \hat{y}_{t,1:k}^{avg}(\theta, \psi) - y_{t,1:k}^{avg} ||_2^2}{k}
 \end{equation}
 and the decoder loss, which is used to optimize the decoder and the embedder modules:
 \begin{equation}
      l^{dec}(\phi, \psi) = \frac{|| \hat{y}_{t+1:t+h}(\phi, \psi) - y_{t+1:t+h} ||_2^2}{h}
 \end{equation}
As backbone networks for the encoder and decoder modules, we used untrained Time-Series Mixer (TSMixer) models \cite{chen2023tsmixer}\footnote{\href{https://github.com/ditschuk/pytorch-tsmixer}{https://github.com/ditschuk/pytorch-tsmixer}}, due to their proved efficiency at extracting temporal and cross-variate patterns stacking simple time-mixing and feature-mixing MLP layers. However, the choice of the backbone network is flexible and not constrained by the architecture.

Further training details are provided in the appendix, in section \ref{sec:training_details}.

\subsection{The dataset}
To validate the proposed architecture, we chose the challenging M5 dataset: the real-world, large-scale dataset used in the M5 forecasting competition \cite{MAKRIDAKIS20221346}. The dataset comprises 30,490 time series representing the number of retail sales of products sold in ten US stores over a period of six years. The features we used, the cardinality of the categorical variables, and their corresponding number of embedding dimensions are provided in the appendix in Table \ref{tab:features_table} and Table \ref{tab:feat_card_table}.

\subsection{Evaluation}
\label{sec:evaluation}
The M5 dataset contains sales data over a period of 1942 days. Following common practice (e.g., see \cite{chen2023tsmixer}), we used the first 1886 days for training, the next 28 days for validation, and the last 28 days for testing. Similarly to \cite{chen2023tsmixer}, we chose a context window $c=35$ and a forecasting horizon $h=28$. As the forecasting horizon includes four weeks, we chose a binning size of a week (i.e., $w=7$ and thus $k=4$). Our main performance measure is the one used by the M5 competition to rank the submissions: the weighted root mean squared scaled error (WRMSSE\footnote{\href{https://github.com/pmrgn/m5-wrmsse}{https://github.com/pmrgn/m5-wrmsse}} — see the appendix for a formal definition). However, we also consider more traditional metrics such as the root mean squared error measured at the fine-grained, daily frequency (RMSEd), and at the coarse-grainded, weekly frequency (RMSEw). Additionally, we considered the RMSE of the residuals (RMSEr), that is, the one of the daily-level time series after subtracting the estimated average over the binning windows. Finally, we also considered the median fraction of explained variance (MFEV) and the mean absolute deviation (MAD).

\subsection{Benchmark models}
\label{sec:benchmark}
Our main comparison is against the monolithic TSMixer architecture (Mono), which was shown to outperform several established forecasting models on popular datasets, including the M5 dataset \citep{chen2023tsmixer}. Critically, for a fair comparison, we kept the number of parameters comparable by splitting the number of hidden units of the monolithic architecture ($n_h=64$) between the encoder and the decoder modules (i.e., $n_e^{enc}=32$ and $n_e^{dec}=32$).
We also compare our architecture (EncDec) against established time series forecasting models such as the Temporal Fusion Transformer (TFT --- \cite{lim2021temporal}) and the Deep Autoregressive Recurrent Network (DeepAR ---\cite{salinas2020deepar}), and against a simple naive model which always forecasts the average over the context window (CtxWindAVG).
\section{Results}
\begin{table}
  \caption{Results. The table contains the performance of the considered models on the test split. For each performance metric, we indicate the value of the best-performing model with a \textbf{bold} font style and the value of the second-best model with an \underline{underlined} font style. Note that the metric values of the models with a "\ddag" superscript are taken from \cite{chen2023tsmixer}. Also note that our models were trained for a maximum of 100 epochs, whereas the models in \cite{chen2023tsmixer} were trained for up to 300 epochs. NBNLL indicates the negative binomial negative log-likelihood loss, while MSE indicates the mean squared error loss. We refer to section \ref{sec:evaluation} for details on the considered metrics and to section \ref{sec:benchmark} for details on the benchmark models.} 
  \label{tab:results_table_all}
  \centering
  \resizebox{\linewidth}{!}{
      \begin{tabular}{llcccccccc}
        \toprule
        \multirow{2}{*}{Model} & \multirow{2}{*}{Loss} & \multicolumn{6}{c}{Metric} \\
        \cmidrule(l){3-8}
        & & WRMSSE ($\downarrow$) & RMSEd ($\downarrow$) & RMSEw ($\downarrow$) & RMSEr ($\downarrow$) & MFEV ($\uparrow$) & MAD ($\downarrow$) \\
        \midrule
        CtxWindAVG &  & 1.085 & 2.242 & 1.316 & 1.815 & 0.000 & 1.066 \\ 
        \midrule 
        DeepAR$^\ddag$ & NBNLL & 0.789 &  &  &  &  &  \\ 
        TFT$^\ddag$ & NBNLL & 0.670 &  &  &  &  &  \\ 
        TSMixer-Ext$^\ddag$ & NBNLL & 0.640 &  &  &  &  &  \\ 
        \midrule 
        MonoNB & NBNLL & 0.681 & 2.179 & 1.269 & 1.771 & 0.024 & 1.053 \\ 
        MonoMSE & MSE & 0.672 & 2.192 & 1.330 & 1.742 & 0.026 & 1.058 \\ 
        \midrule 
        EncDecNB & NBNLL & \underline{0.634} & \textbf{2.091} & \textbf{1.178} & \textbf{1.728} & \underline{0.027} & \textbf{1.035} \\ 
        EncDecMSE & MSE & \textbf{0.620} & \underline{2.146} & \underline{1.263} & \underline{1.735} & \textbf{0.028} & \underline{1.051} \\
      \bottomrule
      \end{tabular}
  }
\end{table}

The results --- summarized in Table \ref{tab:results_table_all} --- indicate that the encoder-decoder architectures (EncDecNB and EncDecMSE) outperform the monolithic ones (MonoNB and MonoMSE) across all the considered metrics. The results also show that the encoder-decoder architectures outperform established forecasting models (DeepAR, TFT, and TSMixer-Ext) in terms of WRMSSE despite being trained for a lower number of epochs (100 vs 300 epochs).

Interestingly, in contrast to previous works \citep{chen2023tsmixer,salinas2020deepar} that reported large performance boost when using the negative binomial negative log-likelihood (NBNLL) loss to train models on demand data characterized by bursty behaviour and widely varying magnitudes (similar to the M5 dataset), we were able to train models equally effectively with both the NBNLL and the MSE losses. As a matter of fact, in our case, we observed that MSE-trained models tend to outperform NBNLL-trained ones on two of the six considered metrics, namely the WRMSSE and MFEV metrics. This seems to suggest that MSE-trained models are better at capturing the behaviour of the time series with larger magnitudes and variances, while, conversely, NBNLL-trained models are better at capturing the behavior of the time series with smaller magnitudes and variances. 

Finally, we note that the encoder-decoder architecture presented in this work was primarily designed to address the challenge of temporal hierarchical forecasting. As such, improved performance at the coarse, weekly level (RMSEw) was an expected outcome of a competent method. However, the general performance boost shown by our architecture suggests that the inductive bias we enforced is conducive to overall improved forecasting accuracy. Particularly noteworthy is, perhaps, the improvement in WRMSSE, which points to better performance in cross-sectional hierarchical forecasting.

\bibliographystyle{plainnat}
\bibliography{main}
\clearpage
\appendix
\section{Appendix}
\begin{table}[]
  \caption{Features table. All the features we provide to the network are listed here. Note that the network receives in input only the $c$ past values of the daily\_sales; for all the other dynamic continuous variables, $c+h$ values are instead provided.}
  \label{tab:features_table}
  \centering
\begin{tabular}{@{}llll@{}}
\toprule
\multicolumn{4}{c}{Features}                                         \\ \midrule
\multicolumn{2}{c}{Static}   & \multicolumn{2}{c}{Dynamic}           \\ \midrule
Categorical    & Continuous  & Categorical      & Continuous         \\ \midrule
item\_id       & avg\_price  & weekday          & daily\_sales ($c$) \\
store\_id      & avg\_nsales & month            & daily\_price       \\
department\_id &             & cultural\_event  & year               \\
category\_id   &             & sporting\_event  &                    \\
state\_id      &             & religious\_event &                    \\
               &             & national\_event  &                    \\
               &             & snap\_day        &                    \\ \bottomrule
\end{tabular}
\end{table}
\begin{table}[]
  \caption{Feature cardinality table. Note that the number of embedding dimensions was determined according to the formula: $n_{emb} = \left\lceil 6r^{0.25} \right\rceil$. The binary variables were not embedded as we did not find this to be beneficial.}
  \label{tab:feat_card_table}
  \centering
  \resizebox{\linewidth}{!}{
\begin{tabular}{@{}cccccc@{}}
\toprule
\multicolumn{6}{c}{Categorical Features}                                                                  \\ \midrule
\multicolumn{3}{c}{Static}                         & \multicolumn{3}{c}{Dynamic}                          \\ \midrule
Name           & Cardinality & Num. Emb. Dim. & Name             & Cardinality & Num. Emb. Dim. \\ \midrule
item\_id       & 3049        & 45                  & weekday          & 7           & 10                  \\
store\_id      & 10          & 11                  & month            & 12          & 12                  \\
department\_id & 7           & 10                  & cultural\_event  & 2           & 1                   \\
category\_id   & 3           & 8                   & sporting\_event  & 2           & 1                   \\
state\_id      & 3           & 8                   & religious\_event & 2           & 1                   \\
               &             &                     & national\_event  & 2           & 1                   \\
               &             &                     & snap\_day        & 2           & 1                   \\ \bottomrule
\end{tabular}
}
\end{table}
\subsubsection{Input features details}
In this section, we provide the input features we used in Table \ref{tab:features_table}. We also provide the cardinality of the categorical variables and their
corresponding number of embedding dimensions in Table \ref{tab:feat_card_table}.
\subsubsection{Training details}
\label{sec:training_details}
The encoder-decoder architecture was trained using two loss functions: the encoder loss $l^{enc}(\theta, \psi)$ --- which is used to compute the gradients of
the encoder $e(z;\theta)$ and the embedder $b(z;\psi)$ --- and the decoder loss $l^{dec}(\phi, \psi)$ --- which is used to compute the gradients of the decoder $d(z;\phi)$ and the
embedder $b(z;\psi)$. More specifically, while the encoder and decoder modules are optimized using their respective loss functions, the embedder is optimized using the sum of these loss functions; that is:
\begin{align}
\nabla_{\theta}^{enc} &:= \nabla_{\theta}l^{enc}(\theta, \psi) \\
\nabla_{\phi}^{dec} &:= \nabla_{\phi}l^{dec}(\phi, \psi) \\
\nabla_{\psi}^{emb} &:= \nabla_{\psi}l^{enc}(\theta, \psi) + \nabla_{\psi}l^{dec}(\phi, \psi)
\end{align}
For each training step, we perform a single forward pass through the model to compute the coarse-grained predictions $\smash{\hat{y}_{t,1:k}^{avg}(\theta,\psi)}$ and the fine-grained ones $\hat{y}_{t \plus 1:t \plus h}^{}(\phi,\psi)$. Importantly, the decoder computes the fine-grained predictions based on a detached copy of the coarse-grained predictions (represented by the "detach" block in Fig.\ref{fig:architecture}). This ensures that the fine-grained predictions, and thus the decoder loss and the decoder gradients, do not depend on the encoder parameters $\theta$.

The training hyperparameters are provided in Table \ref{tab:hyperparameters}. We reused most of the hyperparameters specified in \cite{chen2023tsmixer} and performed a grid search for the learning rate and the loss-rescaling option, which specifies whether to rescale the time series to their original range before computing the loss. All models were trained using the Adam \citep{kingma2014adam} optimizer. The minibatches had size 30 and contained random items sampled by a weighted sampler with weights proportional to the average number of sales \citep{salinas2020deepar}. The best model configurations were selected based on RMSEd computed on the validation set.
\subsubsection{The weighted root mean squared scaled error (WRMSSE)}
The main metric we used to evaluate the performance of our models is the WRMSSE: the metric used to rank the performance of the point forecasts in the M5 Forecasting Accuracy Competition\footnote{\href{https://github.com/Mcompetitions/M5-methods/blob/master/M5-Competitors-Guide.pdf}{https://github.com/Mcompetitions/M5-methods/blob/master/M5-Competitors-Guide.pdf}}. WRMSSE aims to measure the ability of a model to coherently forecast data structured in cross-sectional hierachies such as those in the M5 dataset, which contains 30,490 time series of unit sales that can be grouped in 12 aggregation levels, for a total of 42,840 time series.

Specifically, WRMSSE is defined by:
\begin{equation}
WRMSSE = \sum_{i=1}^{42,840} w_i \times RMSSE_i
\end{equation}
with:
\begin{equation}
RMSSE = \sqrt{
\frac{
\frac{1}{h} \sum_{t=n+1}^{n+h} \left( y_t - \hat{y}_t \right)^2
}{
\frac{1}{n-1} \sum_{t=2}^{n} \left( y_t - y_{t-1} \right)^2
}
}
\end{equation}

In the equation above, $h$ is the forecasting horizon, and $n$ is the number of training days. Thus, RMSSE scales the RMSE of the test predictions with respect to the RMSE of the training predictions of a naive random walk model. Therefore, WRMSSE is a weighted average of RMSSE computed for all the atomic and aggregated time series in the dataset, accurately tracking the ability of a model to make accurate forecasts at each level of the hierarchy. Importantly, the weights $w_i$ are chosen to be proportional to the total sales volume of the atomic and aggregated time series, which is estimated from the sales recorded in the validation window.

\begin{table}[]
\caption{Hyperparameters table. The table reports the hyperparameters of the models shown in Table \ref{tab:results_table_all}. All models were trained for a maximum of 100 epochs, without dropout, and with gradient clipping with a threshold value set to 1.}
\label{tab:hyperparameters}
\centering
\resizebox{\linewidth}{!}{
\begin{tabular}{@{}lclcccc@{}}
\toprule
\multirow{2}{*}{Model} & \multicolumn{6}{c}{Hyperparameters}                                                                                             \\ \cmidrule(l){2-7} 
                       & \multicolumn{1}{l}{Name}         & \multicolumn{1}{c}{Loss} & Loss Rescaling & Learning Rate        & Hidden Units & TSM Blocks \\
\multicolumn{1}{c}{}   & \multicolumn{1}{l}{Search Space} & \multicolumn{1}{c}{}     & \{Yes, No\}    & \{1e-4, 1e-3, 4e-3\} &              &            \\ \midrule
MonoNB                 &                                  & NBNLL                    & Yes            & 1e-3                 & 64           & 2          \\
MonoMSE                &                                  & MSE                      & No             & 1e-3                 & 64           & 2          \\ \midrule
EncDecMB               &                                  & NBNLL                    & No             & 4e-3                 & 32+32        & 2+2        \\
EncDecMSE              &                                  & MSE                      & No             & 1e-4                 & 32+32        & 2+2        \\ \bottomrule
\end{tabular}
}
\end{table}

\mycomment{
The abstract paragraph should be indented \nicefrac{1}{2}~inch (3~picas) on
both the left- and right-hand margins. Use 10~point type, with a vertical
spacing (leading) of 11~points.  The word \textbf{Abstract} must be centered,
bold, and in point size 12. Two line spaces precede the abstract. The abstract
must be limited to one paragraph.
Please read the instructions below carefully and follow them faithfully.

\subsection{Style}

Papers to be submitted to the \textbf{Time Series in the Age of Large Models Workshop} must be prepared according to the
instructions presented here.
These style instructions are adapted from the NeurIPS main conference  \LaTeX{} style.
Papers may only be up to \textbf{four pages long},
including figures.
Additional pages containing \emph{only} acknowledgments, references and optional appendices are allowed.
Papers that exceed the page limit will not be reviewed, or in any other way considered for presentation at the conference.

Authors are required to use the this \LaTeX{} style files obtainable at the
workshop website as indicated below. Please make sure you use the current files
and not previous versions. Tweaking the style files may be grounds for
rejection.

\subsection{Retrieval of style files}

The style files for this workshop and other information are available on
the at
\begin{center}
  \url{https://neurips-time-series-workshop.github.io/}
\end{center}

The \LaTeX{} style file contains three optional arguments: \verb+final+, which
creates a camera-ready copy, \verb+preprint+, which creates a preprint for
submission to, e.g., arXiv, and \verb+nonatbib+, which will not load the
\verb+natbib+ package for you in case of package clash.

\paragraph{Preprint option.}
If you wish to post a preprint of your work online, e.g., on arXiv, using the
NeurIPS style, please use the \verb+preprint+ option. This will create a
nonanonymized version of your work with the text ``Preprint. Work in progress.''
in the footer. This version may be distributed as you see fit. Please \textbf{do
  not} use the \verb+final+ option, which should \textbf{only} be used for
papers accepted to the workshop.

At submission time, please omit the \verb+final+ and \verb+preprint+
options. This will anonymize your submission and add line numbers to aid
review. Please do \emph{not} refer to these line numbers in your paper as they
will be removed during generation of camera-ready copies.

The file \verb+main.tex+ may be used as a ``shell'' for writing your
paper. All you have to do is replace the author, title, abstract, and text of
the paper with your own.

The formatting instructions contained in these style files are summarized in
Sections \ref{gen_inst}, \ref{headings}, and \ref{others} below.

\section{General formatting instructions}
\label{gen_inst}

The text must be confined within a rectangle 5.5~inches (33~picas) wide and
9~inches (54~picas) long. The left margin is 1.5~inch (9~picas).  Use 10~point
type with a vertical spacing (leading) of 11~points.  Times New Roman is the
preferred typeface throughout, and will be selected for you by default.
Paragraphs are separated by \nicefrac{1}{2}~line space (5.5 points), with no
indentation.

The paper title should be 17~point, initial caps/lower case, bold, centered
between two horizontal rules. The top rule should be 4~points thick and the
bottom rule should be 1~point thick. Allow \nicefrac{1}{4}~inch space above and
below the title to rules. All pages should start at 1~inch (6~picas) from the
top of the page.

For the final version, authors' names are set in boldface, and each name is
centered above the corresponding address. The lead author's name is to be listed
first (left-most), and the co-authors' names (if different address) are set to
follow. If there is only one co-author, list both author and co-author side by
side.

Please pay special attention to the instructions in Section \ref{others}
regarding figures, tables, acknowledgments, and references.

\section{Headings: first level}
\label{headings}

All headings should be lower case (except for first word and proper nouns),
flush left, and bold.

First-level headings should be in 12-point type.

\subsection{Headings: second level}

Second-level headings should be in 10-point type.

\subsubsection{Headings: third level}

Third-level headings should be in 10-point type.

\paragraph{Paragraphs}

There is also a \verb+\paragraph+ command available, which sets the heading in
bold, flush left, and inline with the text, with the heading followed by 1\,em
of space.

\section{Citations, figures, tables, references}
\label{others}

These instructions apply to everyone.

\subsection{Citations within the text}

The \verb+natbib+ package will be loaded for you by default.  Citations may be
author/year or numeric, as long as you maintain internal consistency.  As to the
format of the references themselves, any style is acceptable as long as it is
used consistently.

The documentation for \verb+natbib+ may be found at
\begin{center}
  \url{http://mirrors.ctan.org/macros/latex/contrib/natbib/natnotes.pdf}
\end{center}
Of note is the command \verb+\citet+, which produces citations appropriate for
use in inline text.  For example,
\begin{verbatim}
   \citet{hasselmo} investigated\dots
\end{verbatim}
produces
\begin{quote}
  Hasselmo, et al.\ (1995) investigated\dots
\end{quote}

If you wish to load the \verb+natbib+ package with options, you may add the
following before loading the \verb+timeseries_workshop+ package:
\begin{verbatim}
   \PassOptionsToPackage{options}{natbib}
\end{verbatim}

If \verb+natbib+ clashes with another package you load, you can add the optional
argument \verb+nonatbib+ when loading the style file:
\begin{verbatim}
   \usepackage[nonatbib]{timeseries_workshop}
\end{verbatim}

As submission is double blind, refer to your own published work in the third
person. That is, use ``In the previous work of Jones et al.\ [4],'' not ``In our
previous work [4].'' If you cite your other papers that are not widely available
(e.g., a journal paper under review), use anonymous author names in the
citation, e.g., an author of the form ``A.\ Anonymous.''

\subsection{Footnotes}

Footnotes should be used sparingly.  If you do require a footnote, indicate
footnotes with a number\footnote{Sample of the first footnote.} in the
text. Place the footnotes at the bottom of the page on which they appear.
Precede the footnote with a horizontal rule of 2~inches (12~picas).

Note that footnotes are properly typeset \emph{after} punctuation
marks.\footnote{As in this example.}

\subsection{Figures}

\begin{figure}
  \centering
  \fbox{\rule[-.5cm]{0cm}{4cm} \rule[-.5cm]{4cm}{0cm}}
  \caption{Sample figure caption.}
\end{figure}

All artwork must be neat, clean, and legible. Lines should be dark enough for
purposes of reproduction. The figure number and caption always appear after the
figure. Place one line space before the figure caption and one line space after
the figure. The figure caption should be lower case (except for first word and
proper nouns); figures are numbered consecutively.

You may use color figures.  However, it is best for the figure captions and the
paper body to be legible if the paper is printed in either black/white or in
color.

\subsection{Tables}

All tables must be centered, neat, clean and legible.  The table number and
title always appear before the table.  See Table~\ref{sample-table}.

Place one line space before the table title, one line space after the
table title, and one line space after the table. The table title must
be lower case (except for first word and proper nouns); tables are
numbered consecutively.

Note that publication-quality tables \emph{do not contain vertical rules.} We
strongly suggest the use of the \verb+booktabs+ package, which allows for
typesetting high-quality, professional tables:
\begin{center}
  \url{https://www.ctan.org/pkg/booktabs}
\end{center}
This package was used to typeset Table~\ref{sample-table}.

\begin{table}
  \caption{Sample table title}
  \label{sample-table}
  \centering
  \begin{tabular}{lll}
    \toprule
    \multicolumn{2}{c}{Part}                   \\
    \cmidrule(r){1-2}
    Name     & Description     & Size ($\mu$m) \\
    \midrule
    Dendrite & Input terminal  & $\sim$100     \\
    Axon     & Output terminal & $\sim$10      \\
    Soma     & Cell body       & up to $10^6$  \\
    \bottomrule
  \end{tabular}
\end{table}

\section{Final instructions}

Do not change any aspects of the formatting parameters in the style files.  In
particular, do not modify the width or length of the rectangle the text should
fit into, and do not change font sizes (except perhaps in the
\textbf{References} section; see below). Please note that pages should be
numbered.

\section{Preparing PDF files}

Please prepare submission files with paper size ``US Letter,'' and not, for
example, ``A4.''

Fonts were the main cause of problems in the past years. Your PDF file must only
contain Type 1 or Embedded TrueType fonts. Here are a few instructions to
achieve this.

\begin{itemize}

\item You should directly generate PDF files using \verb+pdflatex+.

\item You can check which fonts a PDF files uses.  In Acrobat Reader, select the
  menu Files$>$Document Properties$>$Fonts and select Show All Fonts. You can
  also use the program \verb+pdffonts+ which comes with \verb+xpdf+ and is
  available out-of-the-box on most Linux machines.

\item The IEEE has recommendations for generating PDF files whose fonts are also
  acceptable for this workshop. Please see
  \url{http://www.emfield.org/icuwb2010/downloads/IEEE-PDF-SpecV32.pdf}

\item \verb+xfig+ "patterned" shapes are implemented with bitmap fonts.  Use
  "solid" shapes instead.

\item The \verb+\bbold+ package almost always uses bitmap fonts.  You should use
  the equivalent AMS Fonts:
\begin{verbatim}
   \usepackage{amsfonts}
\end{verbatim}
followed by, e.g., \verb+\mathbb{R}+, \verb+\mathbb{N}+, or \verb+\mathbb{C}+
for $\mathbb{R}$, $\mathbb{N}$ or $\mathbb{C}$.  You can also use the following
workaround for reals, natural and complex:
\begin{verbatim}
   \newcommand{\RR}{I\!\!R} %real numbers
   \newcommand{\Nat}{I\!\!N} %natural numbers
   \newcommand{\CC}{I\!\!\!\!C} %complex numbers
\end{verbatim}
Note that \verb+amsfonts+ is automatically loaded by the \verb+amssymb+ package.

\end{itemize}

If your file contains type 3 fonts or non embedded TrueType fonts, we will ask
you to fix it.

\subsection{Margins in \LaTeX{}}

Most of the margin problems come from figures positioned by hand using
\verb+\special+ or other commands. We suggest using the command
\verb+\includegraphics+ from the \verb+graphicx+ package. Always specify the
figure width as a multiple of the line width as in the example below:
\begin{verbatim}
   \usepackage[pdftex]{graphicx} ...
\end{verbatim}
See Section 4.4 in the graphics bundle documentation
(\url{http://mirrors.ctan.org/macros/latex/required/graphics/grfguide.pdf})

A number of width problems arise when \LaTeX{} cannot properly hyphenate a
line. Please give LaTeX hyphenation hints using the \verb+\-+ command when
necessary.

\begin{ack}
Use unnumbered first level headings for the acknowledgments. All acknowledgments
go at the end of the paper before the list of references. Moreover, you are required to declare
funding (financial activities supporting the submitted work) and competing interests (related financial activities outside the submitted work).
More information about this disclosure can be found at: \url{https://neurips.cc/Conferences/2022/PaperInformation/FundingDisclosure}.

Do {\bf not} include this section in the anonymized submission, only in the final paper. You can use the \texttt{ack} environment provided in the style file to autmoatically hide this section in the anonymized submission.
\end{ack}

\section*{References}

References follow the acknowledgments. Use unnumbered first-level heading for
the references. Any choice of citation style is acceptable as long as you are
consistent. It is permissible to reduce the font size to \verb+small+ (9 point)
when listing the references.
Note that the Reference section does not count towards the page limit.
\medskip

{
\small

[1] Alexander, J.A.\ \& Mozer, M.C.\ (1995) Template-based algorithms for
connectionist rule extraction. In G.\ Tesauro, D.S.\ Touretzky and T.K.\ Leen
(eds.), {\it Advances in Neural Information Processing Systems 7},
pp.\ 609--616. Cambridge, MA: MIT Press.

[2] Bower, J.M.\ \& Beeman, D.\ (1995) {\it The Book of GENESIS: Exploring
  Realistic Neural Models with the GEneral NEural SImulation System.}  New York:
TELOS/Springer--Verlag.

[3] Hasselmo, M.E., Schnell, E.\ \& Barkai, E.\ (1995) Dynamics of learning and
recall at excitatory recurrent synapses and cholinergic modulation in rat
hippocampal region CA3. {\it Journal of Neuroscience} {\bf 15}(7):5249-5262.
}





}

\end{document}